\documentclass[a4paper]{new_tlp}

\newif\ifdraft\drafttrue
\newif\ifinlineref\inlinereffalse
\newif\iffinal\finalfalse
\newif\ifextended\extendedfalse
\newif\ifdotikz\dotikzfalse
\newif\ifmakeallproofsinline\makeallproofsinlinefalse

\newif\iftplp\tplpfalse
\tplpfalse

\inlinereftrue
\draftfalse
\finaltrue
\iftplp
\extendedfalse
\else
\extendedtrue
\fi

\usepackage{times}
\usepackage[scaled]{helvet}
\usepackage{courier}

\usepackage{latexsym}
\usepackage{amsmath}
\usepackage{amssymb}
\usepackage{harpoon}
\usepackage{multirow}
\usepackage{paralist}
\usepackage{listings}
\usepackage{etex}

\usepackage{url}
\makeatletter
\let\UrlSpecialsOld\UrlSpecials
\def\UrlSpecials{\UrlSpecialsOld\do\/{\Url@slash}\do\_{\Url@underscore}}%
\def\Url@slash{\@ifnextchar/{\kern-.11em\mathchar47\kern-.2em}%
    {\kern-.0em\mathchar47\kern-.08em\penalty\UrlBigBreakPenalty}}
\def\Url@underscore{\nfss@text{\leavevmode \kern.06em\vbox{\hrule\@width.3em}}}
\makeatother

\usepackage{color}

\usepackage{tikz}
\pgfrealjobname{dlvhexprogressreport}
\usetikzlibrary{shapes,arrows,backgrounds,chains,%
matrix,patterns,arrows,decorations.pathmorphing,decorations.pathreplacing,%
positioning,fit,calc,decorations.text,shadows,plotmarks%
}

\iffinal\else
\fi
\ifdraft
\usepackage{comments}
\else
\newcommand{\comment}[1]{}
\fi

\usepackage{paralist}
\usepackage{enumerate}
\usepackage{booktabs}
\usepackage{alltt}
\usepackage{caption}

\usepackage{graphicx}
\graphicspath{{figures/}}
\usepackage{subfig}
\usepackage{ifpdf}
\ifpdf
\usepackage{microtype}
\fi

\ifdraft
\usepackage{pdfsync}
\usepackage{srcltx}
\else
\fi

\ifdotikz
\usepackage{tikz}
\pgfrealjobname{hexcdcl}
\usetikzlibrary{shapes,arrows,backgrounds,chains,%
matrix,patterns,arrows,decorations.pathmorphing,decorations.pathreplacing,%
positioning,fit,calc,decorations.text,shadows%
}
\else
\long\def\beginpgfgraphicnamed#1#2\endpgfgraphicnamed{\includegraphics{#1}}
\fi

\ifmakeallproofsinline
\newcommand{\myinlineproof}[1]{#1}
\newcommand{\mylocatedproof}[1]{}
\else
\newcommand{\myinlineproof}[1]{}
\newcommand{\mylocatedproof}[1]{#1}
\fi

\usepackage[vlined,algoruled]{algorithm2e}
\SetAlFnt{\sffamily\tiny}

\newcommand{\leanparagraph}[1]{\smallskip\noindent\textbf{#1}. }

\renewcommand{\vec}[1]{{\bf #1}}

\newcommand{\ext}[3]{\ensuremath{\amp{#1}[#2](#3)}}
\DeclareMathOperator{\naf}{not}

\newcommand{\extfun}[1]{\ensuremath{f_{\text{\sl\&}#1}}}

\newcommand{\extsem}[4]{\ensuremath{f_{\text{\sl\&}#1}(#2,#3,#4)}}

\newcommand{\amp}[1]{\ensuremath{\text{\textsl{{\&}}}\!\mathit{#1}}}

\newcommand{\nop}[1]{}

\newcommand\hex{{\sc hex}}
\newcommand\dlvhex{{\sc dlvhex}}

\newcommand{\dlv}[0]{\texttt{DLV}}
\newcommand{\clasp}[0]{{\sc clasp}}

\newcommand{\smodels}[0]{{\sc smodels}}

\newcommand{\T}{\mathbf{T}}
\newcommand{\F}{\mathbf{F}}
\newcommand{\U}{\mathbf{U}}

\newcommand{\Assignment}{\ensuremath{\mathbf{A}}}

\newcommand{\Program}{\ensuremath{\Pi}}

\newcommand{\BIGOP}[1]
{\mathop{\mathchoice%
{\raise-0.22em\hbox{\huge $#1$}}%
{\raise-0.05em\hbox{\Large $#1$}}{\hbox{\large $#1$}}{#1}}}

{\end{list}}

\usepackage{ntheorem}

\newtheorem{example}{Example}

{\theorembodyfont{\normalfont}
\theoremheaderfont{\itshape}
\newtheorem*{proofsketch}{Proof (Sketch).}
}

\newcommand{\gringo}{{\sc gringo}}

  \title[Theory and Practice of Logic Programming]
        {The {\sc dlvhex} System for Knowledge Representation: Recent Advances (System Description)
		\iffinal%
		\thanks{This research has been supported by the Austrian Science
			Fund (FWF) project P27730.}
		\fi
		}
	
  \author[C.~Redl]
		{Christoph Redl\\
		Institut f\"ur Informationssysteme, Technische Universit\"at Wien\\
		\iffinal%
		Favoritenstra\ss{}e\ 9-11, A-1040 Vienna, Austria\\
		\fi
		\email{redl@kr.tuwien.ac.at}}

\begin{document}

\maketitle

\begin{abstract}
The \dlvhex{} system implements the \hex-semantics,
which integrates answer set programming (ASP) with arbitrary external sources.
Since its first release ten years ago, significant advancements were achieved.
Most importantly, the exploitation of properties of external sources led to
efficiency improvements and flexibility enhancements of the language,
and technical improvements on the system side increased user's convenience.
In this paper, we present the current status of the system
and point out the most important recent enhancements over early versions.
While existing literature focuses on theoretical aspects and specific components,
a bird's eye view of the overall system is missing.
In order to promote the system for real-world applications,
we further present applications which were already successfully realized on top of \dlvhex{}.
\iftplp
\else

This paper is under consideration for acceptance in Theory and Practice of Logic Programming.
\fi
\end{abstract}  

\iffinal
\fi

\renewcommand{\url}[1]{#1}

\section{Introduction}
\label{sec:introduction}

	Answer Set Programming (ASP) is a declarative programming approach
	which has been gaining popularity for many
	applications in artificial intelligence and beyond~\cite{brew-etal-11-asp}.
	Features such as the use of variables as a shortcut for all ground instances,
	aggregates and optimization statements, distinguish ASP from SAT and
	simplify the process of problem solving in many cases.	
	However, since not all data or computation sources can (easily and effectively) be
	encoded in an ASP program, extensions of the formalism towards the integration of
	other formalisms are needed.
	
	To this end, \hex{}-programs extend ASP with arbitrary external sources (which are realized in C++ or Python)
	by the use of so-called \emph{external atoms}.
	Intuitively, the logic program sends information, given by constants and/or predicate extensions, to the external source,
	which returns output values that are imported into the program.
	For instance, the external atom $\ext{\mathit{synonym}}{\mathit{metro}}{X}$
	might be used to find the synonyms $X$ of $\mathit{metro}$, e.g. $\mathit{subway}$ and $\mathit{tube}$.
	Notably, external atoms can be nonmonotonic, introduce new values which are not part of the program (\emph{value invention}),
	and can be used in recursive rules.
	The generality of external sources is in contrast to previous and dedicated formalisms such as DL-programs~\cite{elst2004}
	or constraint ASP~\cite{DBLP:journals/corr/abs-1210-2287}, which integrate ASP only with a concrete
	other formalism. \hex-programs subsume these other formalisms and also well-known ASP extensions such as aggregates.

	However, expressiveness of a formalism alone it not sufficient. Instead, also an efficient and convenient implementation is needed
	to attract users; recall that also the success of ASP depends considerably on expressive, efficient and easy-to-use systems like
	\clasp{}~\cite{gkkoss2011-aicomm}\footnote{\url{http://potassco.sourceforge.net}},
	\dlv{}~\cite{Leone2006}\footnote{\url{http://www.dlvsystem.com}}, and
	\smodels{}~\cite{simo-etal-2002}\footnote{\url{http://www.tcs.hut.fi/Software/smodels}}.
	The \hex-semantics was implemented in the \dlvhex{} system~\cite{eist2006b}\footnote{\url{http://www.kr.tuwien.ac.at/research/systems/dlvhex}}
	on top of \gringo{} and \clasp{}~\cite{gkkoss2011-aicomm}.
	The system celebrates its 10th anniversary this year and was released in version 2.5.0 earlier this year. 
	While early versions were mainly intended to be used for
	experimental purposes, only in the last three years much effort was spent on turning \dlvhex{} into a
	system for KR tasks which can conveniently be used beyond experimental purposes.
	To this end, we have overcome former limitations of the system
	which prevented its application in practice, including former efficiency problems, restrictions of the language, and technical limitations on the system side.

	In this paper, we report about this progress.
	While some (but not all) of the enhancements discussed in the following
	have already been presented in dedicated works, this was from an algorithmic perspective
	and with focus on specific subproblems which occur when evaluating a \hex-program.
	In contrast, this paper provides a bird's eye view of the system from the user's perspective,
	which is missing so far. 

	After briefly recalling \hex-programs in Section~\ref{sec:preliminaries},
	we present the novelties compared to earlier versions of the system.
	We group the enhancements in two main sections:
	\begin{itemize}
		\item Section~\ref{sec:externalsourceproperties} presents enhancements based on the \textbf{exploitation of known properties of external sources}
			such as monotonicity or functionality.
			We first discuss the types of properties supported by the system and show how they are specified (Section~\ref{sec:propertytags}).
			Afterwards, we give an overview about how they are used within the system.
			To this end, we present the two main features based on them, namely \emph{scalability boosts} by advanced learning techniques (Section~\ref{sec:scalability})
			and \emph{language flexibility} due to reduced syntactic limitations (Section~\ref{sec:syntacticlimitations}).
		\item Section~\ref{sec:userconvenience} presents recent extensions towards \textbf{usability and system features}.
			This includes a novel convenient programming interface for providers of external sources (Section~\ref{sec:programminginterface}),
			the integration of support for popular ASP extensions and interoperability (Section~\ref{sec:extensions}),
			and a new dissemination strategy which respects previous user feedback.
	\end{itemize}
	Afterwards, we give an overview about existing applications based on \hex-programs in Section~\ref{sec:applications}
	and discuss how they can benefit from the system improvements.
	We conclude in Section~\ref{sec:discussion}.

	
\section{\hex-Programs}
\label{sec:preliminaries}

	\hex{}-programs~\cite{eist2005} are
	a generalization of
	(disjunctive) extended logic programs under the answer set
	semantics~\cite{gelf-lifs-91} with external atoms.
	Besides \emph{ordinary atoms}\/ of the 
	form $p(\vec{t})$,
	where $p$ is a predicate
	and $\vec{t} = t_1, \dotsc, t_\ell$ is a list of terms (such as strings, integers, symbolic constants, nested function terms),
	rule bodies may also contain
	\emph{external atoms} of the form
	$\ext{g}{\vec{X}}{\vec{Y}}$, 
	where $\amp{g}$ is an external predicate, $\vec{X} = X_1,\ldots, X_l$
	and each $X_i$ 
	is an \emph{input parameter} (which can be either a constant or variable term, or a predicate),
	and $\vec{Y} = Y_1, \ldots, Y_k$ and each $Y_i$ is an \emph{output term}.

	\leanparagraph{Syntax}
	A \emph{\hex-program} (or \emph{program}) consists of rules $r$ of form
	\begin{equation}
	\label{eq:rule}
		a_1\lor\cdots\lor a_h \leftarrow b_1,\dotsc, b_m, \naf\, b_{m+1}, \dotsc, \naf\, b_n \ ,
	\end{equation}
	where each $a_i$ is an (ordinary) atom and
	each~$b_j$ is either an ordinary atom or an external atom, 
	and $h+n>0$; for such a rule $r$ let $B(r) = \{ b_1,\dotsc, b_m, \naf\, b_{m+1}, \dotsc, \naf\, b_n \}$
	denote its \emph{body}.

	\leanparagraph{Semantics}
	An \emph{assignment} $\Assignment$
	is a consistent set of literals of form $\T a$ or $\F a$, where $a$ is an atom which is said to be \emph{true} in $\Assignment$ if $\T a \in \Assignment$, \emph{false} if $\F a \in \Assignment$,
	and \emph{undefined} otherwise. We say that $\Assignment$ is \emph{complete} over a program $\Program$ if for all atoms $a$ in $\Program$ we have either $\T a \in \Assignment$ or $\F a \in \Assignment$.

	The semantics of a \hex-program $\Program$ is defined via
	its grounding $\mathit{grnd}(\Program)$ over a Herbrand universe of constants $\mathcal{C}$ as usual,
	where $\mathcal{C}$ can contain constants which are not in the program and might even be infinite.
	The value of a ground external atom $\ext{g}{\vec{p}}{\vec{c}}$
	wrt.~an assignment $\Assignment$ is given
	by the value
	$\extsem{g}{\Assignment}{\vec{p}}{\vec{c}}$
	of a decidable $1{+}k{+}l$-ary three-valued
	\emph{oracle function} $\extfun{g}$, where $k$ and $l$ are the lengths of $\vec{p}$ and $\vec{c}$, respectively\footnote{In previous works, oracle functions were two-valued;
		we come back to this extension~\cite{ekrw-ijcai16} in Section~\ref{sec:externalsourceproperties}.}.
	The oracle function evaluates to true, false or unknown ($\T$, $\F$ or $\U$), where we assume that
	(i) it evaluates to true or false if $\Assignment$ is complete over $\Program$, and (ii) we have $\extsem{g}{\Assignment'}{\vec{p}}{\vec{c}} = \extsem{g}{\Assignment}{\vec{p}}{\vec{c}}$
	whenever $\Assignment' \supseteq \Assignment$ and $\extsem{g}{\Assignment}{\vec{p}}{\vec{c}} \in \{ \T, \F \}$, i.e.,
	evaluations to true or false do not change when the assignment becomes more complete; we call this property \emph{knowledge-monotonicity}.
	In practice, one often abstracts from the Boolean view and sees an external predicate with input list $\amp{g}[\vec{p}]$ as \emph{computation} of output values $\vec{c}$,
	i.e., determining all values $\vec{c}$ such that $\extsem{g}{\Assignment}{\vec{p}}{\vec{c}} = \T$.

	We note that the definition of the oracle function for
	assignments which are not complete is only for efficiency improvement, as explained in detail in Section~\ref{sec:scalability}. For user's convenience and for backwards compatibility,
	it is also possible to use a two-valued (Boolean) oracle function which is only defined over complete assignments. It is then implicitly assumed to evaluate to unknown for all assignments
	which are not complete.
	
	For (a set of) ground literals, rules, programs, etc., say $O$,
	satisfaction wrt.~a complete assignment~$\Assignment$
	extends naturally from ASP to \hex{}-programs, by
	taking external atoms into account.
	Satisfaction of $O$ under $\Assignment$ is denoted by $\Assignment \models O$.
	In this case we say that $\Assignment$ is a \emph{model} of $O$.
	
	An \emph{answer set} of a \hex-program $\Program$ is a model
	$\Assignment$ of the \emph{FLP-reduct}\footnote{The FLP-reduct is equivalent to the traditional
	reduct for ordinary logic programs
	\cite{gelf-lifs-91}, but more attractive for extensions such as
	aggregates or external atoms.} $f \Program^\Assignment$ of
	$\Program$ wrt.~$\Assignment$, given by $f \Program^\Assignment = \{ r \in \mathit{grnd}(\Program) \mid \Assignment \models B(r)\}$~\cite{flp2011-ai},
	which is subset-minimal, i.e., there exists no model $\Assignment'$ of
	$f \Program^\Assignment$ s.t.~$\{ \T a \in \Assignment'\} \subsetneq \{ \T a \in \Assignment \}$.

	Technically, external atoms are realized as \emph{plugins} of the reasoner using a \emph{programming interface}.
	To this end, the provider of an external source basically implements its oracle function.
	
	\begin{example}
		\label{ex:intro-answer-sets}
		Consider the program
		\begin{equation*}
			\Program {=} \left\{  
			\begin{array}{@{\,}l@{\colon}l@{~~~}l@{\colon}l@{~}l@{}}
				r_1 & \mathit{start}(s). \\
				r_2 & \mathit{scc}(X) \leftarrow \mathit{start}(X). & r_3 & \mathit{scc}(Y) &\leftarrow \mathit{scc}(X), \ext{\mathit{edge}}{X}{Y}. \\[1ex]
			\end{array}\right\}
		\end{equation*}
		where
		$r_1$ selects a node $s$ from an externally defined (finite) graph,
		and $r_2$ and $r_3$ recursively compute the strongly connected component of $s$.
		To this end, the external atom $\ext{\mathit{edge}}{X}{Y}$ is used,
		which is true if $Y$ is directly reachable from $X$ (and false otherwise).
		
		The implementation of $\ext{\mathit{edge}}{X}{Y}$ may look as follows (API details follow in Section~\ref{sec:programminginterface}):
		\begin{lstlisting}[language=Python,basicstyle=\scriptsize]
def edge(x):
  graph=((1,2),(1,3),(2,3))      # simplified implementation; real ones may read a DOT file
  for edge in graph:             # search for outgoing edges of node x
    if edge[0]==x.intValue():
      dlvhex.output((edge[1],))  # output edge target
		\end{lstlisting}
	\end{example}

\section{Exploiting External Source Properties}
\label{sec:externalsourceproperties}

	External sources were seen as black boxes in earlier versions of \dlvhex{}. It was assumed that the system does not have any information about them,
	except that there is an oracle function which decides satisfaction of an external atom under a complete assignment.
	As a consequence, the room for optimizations in the algorithms was limited because the value of an external atom under one assignment
	did not allow for drawing any conclusions about its behavior under other assignments.
	
	However,
	in many practical applications the provider of an external source and/or the \hex-programmer have additional knowledge about the behavior of the source,
	for instance, that the source is monotonic, functional, has a limited domain, returns only elements which are smaller than the input (according to some ordering), etc.
	Knowing such properties allows for implementing more specialized algorithms which are tailored to the particular external sources used in a program.
	We therefore identified a set of \emph{properties} that external sources might have, and allow the user to specify the ones which are fulfilled by a concrete external source.

	\begin{example}
		Suppose $\ext{\mathit{tail}}{X}{Y}$ is true whenever $Y$ is the string which results from string $X$ if the first character is dropped.
		Then the output is always smaller than the input wrt.~string length. 
	\end{example}
	
	The system exploits these properties automatically, mainly for two purposes: in the \emph{learning algorithms for scalability enhancements} and
	in the \emph{grounding component for more flexibility of the language} due to reduced syntactic limitations;
	we discuss these two aspects in more detail in Sections~\ref{sec:scalability} and~\ref{sec:syntacticlimitations}, respectively.
	In addition, there are several other system components which exploit the properties to further speed up the evaluation,
	such as
	skipping various checks if their result is definite due to known behavior of external sources,
	partitioning a reasoning task into smaller independent tasks,
	avoiding unnecessary evaluations of external atoms, and drawing deterministic conclusions rather than guessing.

	However, as this paper presents the system from user's perspective, we focus on \emph{which} properties can be specified, \emph{how} the user can do that,
	and give a rough idea of how the system makes use of this information,
	but we refrain from discussing the involved algorithms in detail.
	This is in line with the goal of these properties: the user can benefit from the advantages when specifying them,
	but without the need to care about how the system is going to exploit this information. Instead,
	the user can generally expect that the more information is available to the system,
	the more efficient evaluation will be; if the added information does not yield a speedup, it does at least no harm.\footnote{
		The only property related to potential performance decrease is provision of a \emph{three-valued semantics} as
		additional calls of the external source are sometimes counterproductive~\cite{ekrw-ijcai16}.
		However, even then the property itself does not harm since it is only exploited by certain (non-default) evaluation heuristics selected via command-line options.
	}
	Some of the properties, such as monotonicity, do even lead to a drop of complexity from $\Sigma^P_2$ to $\mathit{NP}$ for answer set existence checking
	over ground disjunction-free programs, provided that external sources are polynomial~\cite{flp2011-ai}.

	Furthermore, properties also serve as \emph{assertions}:
	if the reasoner observes a behavior of external sources which contradicts the declared properties, appropriate error messages are printed.
	
	\subsection{Specifying Properties}
	\label{sec:propertytags}

		The specification of properties is supported in two ways. The first option is to declare them as part of the external source implementation
		via the \emph{external source interface}. The second option is to specify them as part of the \hex-program using so-called \emph{property tags}.
	
		\leanparagraph{Specification via the External Source Interface}
		Properties are mostly specified via the (C++ or Python) programming interface for external sources.
		To this end, the procedural code which implements external atoms calls specific \emph{setter methods}
		provided by the programming interface
		to inform the system that the source has certain properties.
		
		\begin{example}
			The implementation of a hash function $\ext{\mathit{md5}}{X}{Y}$
			which computes for a string $X$ its MD5 hash value $Y$ might call {\tt prop.setFunctionality(true)}
			to let \dlvhex{} know that for each $X$ there is exactly one $Y$.
			This allows the system, for instance, to conclude that $\ext{\mathit{md5}}{x}{y_2}$ is false without evaluating the external source,
			if it has already found a value $y_1 \not= y_2$ such that $\ext{\mathit{md5}}{x}{y_1}$ is true.
		\end{example}
		
		If a property is declared in this way, the external source is meant to \emph{always} provide a certain behavior, independent of its usage in a certain \hex-program,
		like in case of the computation of a hash value.
		Another example is $\ext{\mathit{diff}}{p,q}{X}$, which computes all values $X$ which are in the extension of $p$ but not in that of $q$ wrt.~assignment $\Assignment$
		(formally, these are all values $x$ such that $\extsem{\mathit{diff}}{\Assignment}{p,q}{x} = \T$).
		This external atom it is always monotone/antimonotone in the first/second parameter,
		which can be specified by calling {\tt prop.addMonotonicInputPredicate(0)} and {\tt prop.addAntimonotonicInputPredicate(1)} (cf.~Example~\ref{ex:python}).
		
		\leanparagraph{Specification via Property Tags}
		However, it might also be the case that only a specific usage of an external source in a concrete program has a property.
		Then the implementer of the external source cannot declare it yet;
		instead, only the implementer of the \hex-program has sufficient knowledge and can declare the property as part of
		an external atom in the program.
		
		\begin{example}
			Suppose $\ext{\mathit{greaterThan}}{p, 10}{}$ checks if the sum of integer values $c$ s.t.~$p(c)$ is true is greater than $10$.
			It is not monotone in general if negative integers are allowed, but it is monotone if a program uses only positive integers.
			While the provider of the external source cannot assert the property, the user of the external source in a concrete program,
			who knows the context, can.
		\end{example}
		
		To this end, the \hex{} language and implementation were extended such that external atoms can be followed by \emph{property tags}
		of form $\langle \mathit{list\ of\ properties} \rangle$,
		where the list of properties is comma-separated. Each property is then a whitespace-separated list of constants,
		consisting of a \emph{property type} (first element in the list),
		and a number of \emph{property parameters} (remaining elements in the list), whose number depends on the property type and may also have default values.
		For example, $\ext{\mathit{diff}}{p,q}{X}\langle \mathit{monotonic\ p}, \mathit{antimonotonic\ q}\rangle$
		specifies two properties which
		declare that the external atom is monotonic in $p$ and antimonotonic in $q$ wrt.~their extension in the input assignment. Here, the first property $\mathit{monotonic\ p}$
		uses the property type $\emph{monotonic}$ and the property parameter $p$,
		while the second property $\mathit{antimonotonic\ q}$
		uses the property type $\emph{antimonotonic}$ and the property parameter $q$.
		Another example is
		$\ext{\mathit{greaterThan}}{p, 10}{}\langle monotonic \rangle$,
		which declares that the external source is monotonic in all parameters (because it is monotonic in $p$ and it is trivially monotonic in constant input parameters because they are independent of the input assignment);
		the property type is $\emph{monotonic}$ and no property parameters are explicitly specified,
		which indicates by default that the source is monotonic in all inputs.
		Properties declared by tags are understood to hold \emph{in addition} to those declared via the external source interface
		(stating conflicting properties is not possible with the currently available ones).
		
		\leanparagraph{Supported properties}
		The following list gives an overview about the currently available properties and how to specify them if the property tag language is used
		(but all of them can be specified both via the external source interface or in property tags).
		Each property is explained with an example in order to show the property type and the expected property parameters.
		
		\newcommand{\prop}[4]{\item \textbf{#1}: #4 \\ #3}
			{
			\noindent\begin{itemize}
				\prop{Functionality}{functional}{The external atom adds integers $X$ and $Y$ and is true for their sum $Z$. The source provides exactly one output value for a given input. There are no property parameters.}{$\ext{\mathit{add}}{X,Y}{Z}\langle \mathit{functional} \rangle$}
				\prop{Monotonicity in a parameter}{monotonic p}{The external atom computes the difference of the extensions of $p$ and $q$. The source is monotonic in predicate parameter $p$ (i.e., if the extension of $p$ increases, the output does not shrink), as indicated by the property parameter.}{$\ext{\mathit{diff}}{p,q}{X}\langle \mathit{monotonic\ p} \rangle$}
				\prop{Global monotonicity}{monotonic}{The source computes the set union of the extensions of $p$ and $q$. It is monotonic in all parameters (indicated by the default value of the missing property parameter).}{$\ext{\mathit{union}}{p,q}{X}\langle \mathit{monotonic} \rangle$}
				\prop{Antimonotonicity in a parameter}{antimonotonic p}{The source is antimonotonic in predicate parameter $q$ (i.e., if the extension of $q$ shrinks, the output does not shrink).}{$\ext{\mathit{diff}}{p,q}{X}\langle \mathit{antimonotonic\ q} \rangle$}
				\prop{Global antimonotonicity}{antimonotonic}{The source computes the complement of the extension of $p$ wrt.~a fixed domain. It is antimonotonic in all parameters.}{$\ext{\mathit{complement}}{p}{X}\langle \mathit{antimonotonic} \rangle$}
				\prop{Linearity on atoms}{atomlevellinear}{We have domain independence on the level of atoms, i.e., the source can be separately evaluated for each input atom s.t.~the final result is the union of the results of all evaluations. For instance, the evaluation under assignment $\Assignment = \{ \T p(a), \T p(b), \T q(c) \}$, which yields $\{ a, b, c \}$, can be split up into three evaluations under $\Assignment_1 = \{ \T p(a) \}$, $\Assignment_2 = \{ \T p(b) \}$ and $\Assignment_3 = \{ \T q(c) \}$, which yield $\{a\}$, $\{b\}$ and $\{c\}$, respectively, and their union the result of the evaluation under $\Assignment$. There are no property parameters.}{$\ext{\mathit{union}}{p,q}{X}\langle \mathit{atomlevellinear} \rangle$}
				\prop{Linearity on tuples}{tuplelevellinear}{We have domain independence on the level of tuples in the extensions of predicate input parameters, i.e., the source can be separately evaluated for each pair of atoms $p(\vec{c})$ and $q(\vec{c})$ for all vectors of terms $\vec{c}$ s.t.~the final result is the union of the results of all evaluations. For instance, the evaluation under $\Assignment = \{ \T p(a), \T p(b), \F q(a), \T q(b) \}$, which yields $\{ a \}$, can be split up into two evaluations under $\Assignment_1 = \{ \T p(a), \F q(a) \}$ and $\Assignment_2 = \{ \T p(b), \T q(b) \}$, which yield $\{ a \}$ and $\emptyset$, respectively, and their union in the result of the evaluation under $\Assignment$. However, it would not be correct to split $\Assignment_2$ further up into $\Assignment_{2.1} = \{ \T p(b) \}$ and $\Assignment_{2.2} = \{ \T q(b) \}$ as they would yield the results $\{ b \}$ and $\emptyset$, which would put $b$ into the final result, which differs from the evaluation under $\Assignment$. There are no property parameters.}{$\ext{\mathit{diff}}{p,q}{X}\langle \mathit{tuplelevellinear} \rangle$}
				\prop{Finite domain}{finitedomain o}{Imports the edges of a predefined graph. Both output values can have only finitely many different values. To this end, we specify two properties with type $\mathit{finitedomain}$ with property parameters that identify the output terms $X$ and $Y$ by index ($0$ and $1$, respectively).}{$\ext{\mathit{edges}}{\mathit{graph.dot}}{X,Y}\langle \mathit{finitedomain\ 0}, \mathit{finitedomain\ 1} \rangle$}
				\prop{Finite domain wrt.~the input}{relativefinitedomain i o}{Only constants which already appear in the $0$-th input (indicated by the first property parameter $0$; points in this case to the predicate $p$) may occur as first output term (indicated by the second property parameter $0$). Informally, the difference between sets represented by predicates $p$ and $q$ can only contain elements which appear in the set represented by $p$.}{$\ext{\mathit{diff}}{p,q}{X}\langle \mathit{relativefinitedomain\ 0\ 0} \rangle$}
				\prop{Finite fiber}{finitefiber}{The source computes the square root of $X$. Each element in the output is only produced by finitely many different inputs (in this case, in fact, only by a single input value). There are no property parameters.}{$\ext{\mathit{sqrt}}{X}{Z}\langle \mathit{finitefiber} \rangle$}
				\prop{Well-ordering wrt.~string lengths}{wellorderingstrlen i o}{The source drops the first character of string $X$ and returns the result in $Z$. The $0$-th output (indicated by the second property parameter $0$) is no longer than the longest string in the $0$-th input (indicated by the first property parameter $0$).}{$\ext{\mathit{tail}}{X}{Z}\langle \mathit{wellorderingstrlen\ 0\ 0} \rangle$}
				\prop{General well-ordering}{wellordering i o}{The external atom decrements a given integer. There is an ordering of all constants such that the $0$-th output (second parameter) is no greater than the $0$-th input (first parameter) wrt.~this ordering.}{$\ext{\mathit{decrement}}{X}{Z}\langle \mathit{wellordering\ 0\ 0} \rangle$}
				\prop{Three-valued semantics}{providespartialanswer}{The external source can be evaluated under partial assignments, i.e., it can handle assignments which do not define all atoms, but may evaluate to \emph{undefined} ($\U$) in this case (can be used with any external source if implemented).}{$\ext{\mathit{g}}{\vec{X}}{\vec{Y}}\langle \mathit{providespartialanswer} \rangle$}
			\end{itemize}
			}
		%
		Note that properties are only useful if they are exploited by at least one solving technique or algorithm implemented in the reasoner. It is therefore \emph{not} intended that typical users
		introduce custom properties, but only tag external atoms with existing ones from the above list.
		However, for advanced users who contribute to or customize the reasoner itself, the framework supports easy extension of the parser and data structures.
		Exploiting such a new property in the algorithms might be more sophisticated depending on the particular property and the envisaged goal.

	\subsection{Scalability Boost}
	\label{sec:scalability}

		Traditionally, ground \hex-programs have been evaluated by replacing each external atom $\ext{e}{\vec{p}}{\vec{c}}$
		by an ordinary atom $e_{\amp{e}[\vec{p}]}(\vec{c})$ and introducing a rule $e_{\amp{e}[\vec{p}]}(\vec{c}) \vee \mathit{ne}_{\amp{e}[\vec{p}]}(\vec{c}) \leftarrow$
		to guess its truth value; the resulting program is evaluated by an ordinary ASP solver to produce model candidates.
		Each candidate $\Assignment$ is subsequently checked by testing (i) if the external atom guesses are correct, i.e.,
		if $\Assignment \models e_{\amp{e}[\vec{p}]}(\vec{c})$ iff $\Assignment \models \ext{e}{\vec{\vec{p}}}{\vec{\vec{c}}}$ for all external atoms \ext{e}{\vec{\vec{p}}}{\vec{\vec{c}}},
		and (ii) if assignment $\Assignment$ is a \emph{subset-minimal} model of $f \Program^\Assignment$. If both conditions are satisfied, an answer set has been found.
		However, this approach did not scale well because there are exponentially many independent guesses in the number of external atoms
		in the ground program.
		
		\leanparagraph{Basic approach}
		To overcome the problem, novel evaluation algorithms based on \emph{conflict-driven} techniques have been introduced~\cite{efkr2012-tplp}.
		As in ordinary ASP solving, the input program is translated to a set of \emph{nogoods}, i.e., a set of literals which must not be true at the same time.
		Given this representation, techniques from SAT solving are applied to find an assignment which satisfies all nogoods~\cite{gks2012-aij}.
		Notably, as the encoding as a set of nogoods is of exponential size due to \emph{loop nogoods} which avoid cyclic justifications of atoms, those parts are generated only on-the-fly.
		Moreover, additional nogoods are learned from conflict situations, i.e., violated nogoods which cause the solver to backtrack; this is called \emph{conflict-driven nogood learning}.
		
		The extension of this algorithm towards the integration of external sources into the learning component works as follows.
		Whenever an external atom $\ext{e}{\vec{p}}{\vec{c}}$ is evaluated under an assignment $\Assignment$
		in the checking part (i), the actual truth value under the assignment becomes evident. Then, regardless of whether the guessed value was correct or not,
		one can add a nogood which represents that $e_{\amp{e}[\vec{p}]}(\vec{c})$ must be true under $\Assignment$ if $\Assignment \models \ext{e}{\vec{p}}{\vec{c}}$
		or that $e_{\amp{e}[\vec{p}]}(\vec{c})$ must be false under $\Assignment$ if $\Assignment \not\models \ext{e}{\vec{p}}{\vec{c}}$.
		If the guess was incorrect, the newly learned nogood will trigger backtracking, if the guess was correct,
		the learned nogood will prevent future wrong guesses.
		
		\begin{example}
			As above, suppose $\ext{\mathit{diff}}{p,q}{X}$ computes the set difference between the extensions of predicates $p$ and $q$
			and that it is evaluated under $\Assignment = \{ \T p(a), \T p(b), \F q(a), \T q(b) \}$ with Herbrand universe $\mathcal{C} = \{ a, b \}$.
			Then it can be learned that $\Assignment \models e_{\amp{e}[p,q]}(a)$ by adding the nogood $\{ \T p(a), \T p(b), \F q(a), \T q(b), \F e_{\amp{e}[p,q]}(a) \}$,
			i.e., whenever $p(a), p(b), q(b)$ are true and $q(a)$ is false, then ${\amp{e}[p,q]}(a)$ must not be false.
			Conversely, one can learn that $\Assignment \not\models {\amp{e}[p,q]}(b)$ by adding nogood $\{ \T p(a), \T p(b), \F q(a), \T q(b), \T e_{\amp{e}[p,q]}(b) \}$.
		\end{example}

		Experimental results show a significant, up to exponential speedup~\cite{efkrs2014-jair}. This is explained by
		the exclusion of up to exponentially many guesses by the learned nogoods.
		
		\leanparagraph{Exploiting external source properties}
		The technique was refined by exploiting additional knowledge about external sources in order to keep the learned nogoods small.
		In the previous example, atoms $p(a)$ and $q(a)$ in the assignment are in fact irrelevant when deciding whether $\ext{e}{p,q}{b}$ is true
		because constants $a$ and $b$ are independent (similarly for $p(b)$ and $q(b)$ when deciding $\ext{e}{p,q}{a}$). If this information is available to the system,
		it can be exploited to shrink nogoods to the relevant part such that the search space is pruned more effectively.
		
		One way to gain the required information is to make use of the properties introduced in Section~\ref{sec:propertytags}.
		In particular, the independence of $a$ and $b$ in the previous example can be derived from the property `linearity on tuples'.
		Then the nogood $\{ \T p(a), \T p(b), \F q(a), \T q(b), \F e_{\amp{e}[p,q]}(a) \}$ can be reduced to $\{ \T p(a), \F q(a),$ $\F e_{\amp{e}[p,q]}(a) \}$
		and the nogood $\{ \T p(a), \T p(b), \F q(a), \T q(b),$ $\T e_{\amp{e}[p,q]}(b) \}$ to $\{ \T p(b),$ $\T q(b), \T e_{\amp{e}[p,q]}(b) \}$.
		If monotonicity in $p$ is known in addition, then nogood $\{ \T p(b),$ $\T q(b), \T e_{\amp{e}[p,q]}(b) \}$ can be further simplified to $\{ \T q(b), \T e_{\amp{e}[p,q]}(b) \}$
		by dropping $\T q(a)$
		because ${\amp{e}[p,q]}(b)$ will remain false even if $q(a)$ becomes false.
		
		\leanparagraph{Exploiting three-valued oracle functions}
		Alternatively or in addition to external source properties, also three-valued oracle functions (cf.~Section~\ref{sec:preliminaries}) can be exploited for shrinking learned nogoods
		to the essential part~\cite{ekrw-ijcai16}.		
		If the truth value is already known and will not change when the assignment becomes more complete, then the set of yet unassigned atoms is irrelevant for the output of the external source.
		This is exploited for nogood minimization as follows.
		Whenever a nogood is learned, the system iteratively tries to remove one of the input atoms and evaluate again in order to check if the truth value is still defined.
		If this is the case, the according atom is not necessary and can be removed from the nogood.
		
		For instance, a proper implementation of a three-valued oracle function
		in the previous example allows for reducing $\{ \T p(a), \T p(b), \F q(a), \T q(b), \T e_{\amp{e}[p,q]}(b) \}$
		to $\{ \T q(b), \T e_{\amp{e}[p,q]}(b) \}$ because whenever $\T q(b)$ is in the assignment, it is already definite that $\ext{\mathit{diff}}{p,q}{b}$ is false.

		\leanparagraph{Discussion and Extensions}
		Whether to exploit external source properties, three-valued oracle functions, or both, depends largely on the use case.
		Depending on the type of external source to be realized, the implementation of a three-valued oracle function might be more challenging than of a Boolean one
		(implementing an algorithm which decides over partial assignments is in general more difficult than if all information is known).
		However, it allows for exploiting application-specific knowledge in an optimal way~\cite{ekrw-ijcai16}.
		In contrast, tagging external sources with properties from a list is easy and can still lead to good efficiency.

		
	\subsection{Language Flexibility}
	\label{sec:syntacticlimitations}

		External atoms may introduce constants which do not appear in the program (\emph{value invention}).
		Obviously, this can in general lead to programs which do not have a finite grounding that has the same answer sets as the original program
		(which are defined via the full, possibly infinite grounding $\mathit{grnd}(\Program)$).
		Since this inhibits grounding in general,
		it is crucial to identify classes of programs for which the existence of such a finite grounding is guaranteed;
		we call this property \emph{finite groundability}.
		Traditionally, \emph{strong safety} was used, which basically forbids value invention by recursive external atoms (i.e., external atoms whose input possibly depends on its own output wrt.~the predicate dependency graph,
		for a formal definition cf.~\citeN{eist2006}).
		If only non-recursive external atoms introduce new values, termination is guaranteed.
		However, it turns out that this is only a sufficient but not a necessary criterion, i.e., strong safety is overly restrictive.
		
		\begin{example}
			The program $\Program$ from Example~\ref{ex:intro-answer-sets} is \emph{not} strongly safe because $\ext{\mathit{edge}}{X}{Y}$
			is recursive (output $Y$ may be input to the same external atom by another application of $r_3$) but may introduce values for $Y$ which do not appear in $\Program$.
			However, if one knows that the graph is finite, one can conclude that the recursive introduction of new values will end at some point.
		\end{example}
			
		In the example, the criterion may be circumvented by importing the full domain a priori and adding \emph{domain predicates}, i.e.,
		adding $\mathit{node}(Y)$ to the body of $r_3$ and another rule $\mathit{node}(X) \leftarrow \ext{\mathit{node}}{}{X}$ to import all nodes.
		Then $\ext{\mathit{edge}}{X}{Y}$ does no longer invent values because all possible values for $Y$ are determined in a non-recursive fashion using $\ext{\mathit{node}}{}{X}$.
		However, this comes at the price of importing the whole graph although only a small set of nodes might be in the strongly connected component of $s$.

		Therefore,
		new safety criteria have been introduced which allow for exploiting both syntactic and semantic conditions
		to derive finite groundability, where the latter are based on external source properties as introduced in Section~\ref{sec:propertytags}.
		So-called \emph{liberally safe} \hex-programs are guaranteed to have a finite grounding
		which can be computed using a novel algorithm~\cite{efkr2016-aij}.

		
		\begin{example}
			Let $\ext{\mathit{tail}}{X}{Y}$ drop the first character of string $X$ and return it as $Y$.
			Then $Y$ is no longer than $X$ and
			-- even if used recursively -- it is guaranteed that it can generate only finitely many strings
			because there are only finitely many strings with a length up to the one of $X$.
		\end{example}
		
		In addition to the declaration of predefined properties,
		the generic framework is also extensible such that custom knowledge about external sources can be exploited.
		To this end, providers may implement \emph{safety plugins}, which are integrated into the safety check.
		The safety check itself is fast (at most quadratic in the size of the non-ground program).
		
		The system combines the available information, given by syntactic conditions, specified semantic properties
		and safety plugins in order to check safety of the program.
		This does not only allow for writing programs with fewer syntactic restrictions,
		but the implementation of some applications may be possible in the first place.
		For instance, in \emph{route planning applications}, importing the whole map material a priori
		is practically impossible due to the large amount of data, while a selective import using liberal safety makes the application possible~\cite{efkr2016-aij}.

		In case a program is not safe, the system prints hints such as the rule and the variable for which finiteness during instantiation could not be proven.
		This information is intended to guide the user when providing more information in order to make the program safe, e.g., by adding properties from Section~\ref{sec:propertytags}
		which constrain the values of this variable further.
		Alternatively, a command-line option allows to disable the safety check altogether, in which case there is no guarantee that the reasoner terminates (putting this burden on the user).

\section{Usability and System Features}
\label{sec:userconvenience}

	In this section we present recent work on the system side to improve the user's convenience.
	We start with general remarks on the \dlvhex{} software and its dissemination.
	{\sc dlvhex} was previously only available in source format (released under GNU LGPL) and only for Linux platforms.
	This deployment method turned out to be inconvenient for ASP programmers who want to use the system as is without custom modifications,
	thus we now provide pre-built binaries for all major platforms (Linux-based, OS X and Windows) in addition.
	We further created an online demo of the system under \url{http://www.kr.tuwien.ac.at/research/systems/dlvhex/demo.php}
	which allows for evaluating \hex-programs directly in the browser (the user may specify both the logic program and
	custom Python-implemented external atoms in two input fields). The demo comes with a small set of examples to
	demonstrate the main features of the KR formalism.
	We further provide a manual to support new users of the system~\cite{emrs2015-hexmanual}.

	Next, the following two subsections give an overview of the new Python programming interface and interoperability of the system.

	\subsection{Python Programming Interface}
	\label{sec:programminginterface}

		With earlier versions of the system, users who wanted to integrate custom external sources had to write plugins in C++.
		While this was natural as the reasoner itself is implemented in C++, it was cumbersome and introduced development overhead even
		for experienced developers. This is because multiple configuration, source and header files need to be created even when realizing only a small and simple plugin.
		Also the compilation and linking overhead during development and debugging was considered inconvenient.
		
		As a user-friendly alternative, \dlvhex{} 2.5.0 introduces a plugin API for Python-implemented external sources.
		A plugin consists of a single file (unless the user explicitly wants to use multiple files), which imports a dedicated {\tt dlvhex} package
		and specifies a single method for each external atom.
		Thanks to higher-level features of Python and modern packages, this usually results in much shorter and simpler code than
		with C++-implemented plugins.
		A central {\tt register} method exports the available external atoms and (optionally) their properties
		from Section~\ref{sec:propertytags} to \dlvhex{}.
	
		\begin{example}
			\label{ex:python}
			The following snippet implements $\ext{\mathit{diff}}{p,q}{X}$ for computing the values $X$
			which are in the extension of $p$ but not in that of $q$. It is monotonic in $p$ and antimonotonic in $q$.
			
			\begin{lstlisting}[language=Python,basicstyle=\scriptsize]
import dlvhex

def diff(p,q):
  for x in dlvhex.getTrueInputAtoms():         # for all true input atoms
    if x.tuple()[0] == p:                      # is it of form p(c)?
      if dlvhex.isFalse(dlvhex.storeAtom(      # is the corresponding q(c) false?
                          (q, x.tuple()[1]))):
        dlvhex.output((x.tuple()[1], ));       # then c is in the output

def register():
  prop = dlvhex.ExtSourceProperties()          # inform dlvhex about
  prop.addMonotonicInputPredicate(0)           # monotonicity/antimonotonicity
  prop.addAntimonotonicInputPredicate(1)       # in the first/second parameter
  dlvhex.addAtom("diff", (dlvhex.PREDICATE, dlvhex.PREDICATE), 1, prop)
			\end{lstlisting}
		\end{example}
		
		On the command-line, the call {\tt dlvhex2 --python-plugin=plugin.py prog.hex} loads the external atoms defined in {\tt plugin.py}
		and then evaluates \hex-program {\tt prog.hex}.
		
\begin{figure}[t]
			\centering
			\beginpgfgraphicnamed{figOverview}
			\begin{tikzpicture}[start chain,node distance=0.7cm,every on chain/.style={join=by ->}, every join/.style={line width=1.25pt}]
			\matrix (m) [matrix of nodes, 
			column sep=3mm,
			row sep=3mm,
			nodes={draw, 
			  line width=0.7pt,
			  anchor=center, 
			  text centered,
			  text width=2.75cm
			},
			system/.style={
			  line width=1.25pt,
			  text width=2.8cm, 
			  minimum width=1.5cm, minimum height=9mm
			}
			]
			{
			  |[system]|  Reasoning Component & |[system]| C++ Programming Interface & C++ Plugins \\
			                                           & |[system]| Python Programming Interface & Python Plugins \\
			};

			{ [start chain,every on chain/.style={join}, every join/.style={line width=1.25pt}]
			  \path[line width=0.7pt,<->] (m-1-1) edge node [right] {} (m-1-2);
			  \path[line width=0.7pt,<->] (m-1-2) edge node [right] {} (m-2-2);
			  \path[line width=0.7pt,<->] (m-1-3) edge node [right] {} (m-1-2);
			  \path[line width=0.7pt,<->] (m-2-3) edge node [right] {} (m-2-2);
			};

			\tikzset{my dotted/.style={draw=black, line width=0.5pt,
				dash pattern=on 2pt off 2pt,
				inner sep=2mm, rectangle, rounded corners}};

			\node (dlvhex) [my dotted, fit=(m-1-1) (m-2-2)] {};
			\node at (dlvhex.south) [below, inner sep=2mm] {\dlvhex};
			\end{tikzpicture}
			\endpgfgraphicnamed
			\caption{Architecture of the Python Programming Interface}
			\label{fig:pythoninterface}
		\end{figure}
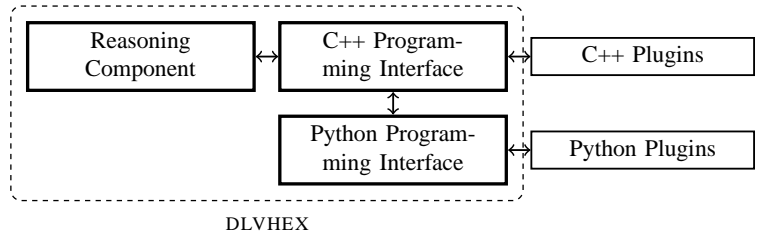

		In the system, the Python programming interface is realized as a wrapper of the generic C++ interface as shown in Figure~\ref{fig:pythoninterface},
		where arcs model both control and data flow. That is, the Python interface uses only the C++ interface but does not communicate
		with the core reasoning components otherwise.
		This turns the Python interface in fact into a special C++ plugin.
		The performance gap between C++ and Python plugins is normally negligible (the update of the Python data structures it in the worst case linear in the number of input atoms),
		unless the plugin is itself computationally expensive.
		Wrappers for other languages can be added similarly and can also be implemented externally, i.e., they do not necessarily need to be part of the \dlvhex{} solver.

		For a complete API description we refer to \url{http://www.kr.tuwien.ac.at/research/systems/dlvhex/}.

	\subsection{ASP-Core-2 Standard, Extensions and Interoperability}
	\label{sec:extensions}

		In the course of the organization of the fourth ASP competition,
		the input language of ASP systems was standardized in the \emph{ASP-Core-2 input language format}~\cite{aspcore2}.
		The {\sc dlvhex} system in its current version supports all features defined in the standard, including
		function symbols, choice rules, conditional literals, aggregates, and weak constraints.
		The supported language is therefore a strict superset of the standard.
		
		The system further supports input and output in CSV format to improve interoperability with other systems such as Unix commands
		or spreadsheet applications. That is, facts may be read from the lines of a CSV file,
		where the different values are mapped to the arguments of a predicate.
		After the computation, the extension of a specified predicate may be written in CSV format
		to allow a seamless further processing by other applications.
		For instance, consider {\tt salary.csv}:%
\begin{verbatim}
joe,smith,2000
sue,johnson,2200
\end{verbatim}%
		It can be read as facts $\mathit{emp}(1, \mathit{joe}, \mathit{smith}, 2000)$ and $\mathit{emp}(2, \mathit{sue}, \mathit{johnson}, 2200)$
		(where the first element is the original line number if relevant) using the \dlvhex{} command-line option {\tt --csvinput=emp,salary.csv}.
		Conversely, results can be output in CSV format.

\section{Applications}
\label{sec:applications}

	We now discuss some applications which were realized on top of \hex-programs.
	In this paper, we focus on applications whose purpose was \emph{not}
	to demonstrate \hex-programs or to evaluate the reasoner. Instead,
	the following applications are motivated by real needs and are interesting by themselves,
	while \hex-programs were merely a means for their realization.
	This witnesses that \hex-programs and \dlvhex{} can be fruitfully applied for real-world problems.

	We discuss the effects of the described system improvements on the applications.
	\iftplp
	\else
	Some quantitative results are included in~\ref{sec:benchmarks}.
	\fi
	However, since this paper gives an overview of the system and not all of the presented improvements are related to efficiency,
	not all of the following applications are suited as performance benchmarks.
	For an extensive empirical evaluation focused on efficiency we refer to~\citeN{efkrs2014-jair} and~\citeN{efkr2016-aij}.

	\leanparagraph{Hybrid planning}
	The application comes from the robotics domain and consists of high-level planning and low-level feasibility checking~\cite{Erdem2016a}.
	High-level plans are sequences of actions towards a goal, while low-level constraints (such as stability issues of robots or intersections of routes) exclude some of the sequences.
	Thus, not all such plans which are possible from high-level perspective are actually executable.
	The separation of the two levels is motivated by the observation that the full integration of low-level constraints
	into the model for high-level planning might blow up the encoding (while it might be feasible in other cases).
	An implementation of hybrid planning on top of \hex-programs was presented, where external atoms are used to perform low-level feasibility checking
	of high-level plans generated in the program.

	\noindent \emph{Effects of improvements:}
	The application uses hand-crafted \emph{custom learning functions} which add custom nogoods during evaluation
	to improve efficiency, cf.~\citeN{Erdem2016a}.
	With the new \dlvhex{} version, three-valued oracle functions can be used instead, which allow for an easier realization of a similar learning behavior.
	Furthermore, the property \emph{tuplelevellinear} can be exploited whenever feasibility checks can be split into multiple independent checks (e.g.~of independent robots),
	and \emph{relativefinitedomain} can be exploited for external atoms used for sensing objects (only objects which appear in the description of the world can be sensed).

	\leanparagraph{Route planning}
	The combination of route planning with side constraints was realized on top of \hex~\cite{efkr2016-aij}.
	An example is planning a tour through multiple locations, where the possibility to get refreshments should be included if the tour is longer than a limit.
	
	\noindent \emph{Effects of improvements:}
	Since the traditional criterion of strong safety disallows recursive value invention,
	previous system versions must import the whole map a priori.
	As this is infeasible for real-world data, the application can in fact only be realized on top of \hex{} by exploiting the improvements from Section~\ref{sec:syntacticlimitations}.
	To this end, \emph{finiteness} of the map used with liberal safety allows for importing only relevant parts of the map
	and solving the problem efficiently~\cite{efkr2016-aij}.

	\leanparagraph{Multi-context Systems}
	Multi-context systems are a framework for integrating heterogeneous knowledge-bases, called \emph{contexts}, which are abstractly identified by sets of belief sets~\cite{be2007}.
	Their integration works via dedicated \emph{bridge rules} which derive information in one context
	based on atoms in other contexts.
	The whole system may become inconsistent although the individual contexts are all consistent. A typical reasoning task is then \emph{inconsistency analysis},
	i.e., the computation of an inconsistency explanation~\cite{efsw2010-kr}, which
	was realized on top of \hex-programs~\cite{befs2010}. The main idea is to realize contexts as external sources.
	Then a \hex-program can access all contexts, compute candidate explanations, and check them against all contexts.
	Experimental results, which demonstrate effectiveness of the learning techniques from Section~\ref{sec:scalability}, can be found in the work by~\citeN{efkrs2014-jair}.

	\noindent \emph{Effects of improvements:}
	Already plain learning (general part of Section~\ref{sec:externalsourceproperties}) is highly effective, cf.~\cite{efkrs2014-jair}.
	The external atoms are \emph{functional}, which does however, since their output is $0$-ary, not lead to additional performance improvements, but does at least not harm.

	\leanparagraph{Complaint management}
	Citizens may raise complaints about issues such as noise or traffic jams as part of e-government.
	A system was realized on top of \dlvhex{}, which ranks complaints by their severity, such that priorities
	can be assigned~\cite{conf/semweb/ZirtilogluY08}.
	While ontologies capture parts of the application, the authors combine them with \hex-programs due to the inherent support
	for nonmonotonic reasoning. This is motivated by the dynamic behavior of complaint management systems, which might need
	to adopt the ranking if new complaints are added.
	
	\noindent \emph{Effects of improvements:}
	The encoding makes use of recursive rules over external atoms, thus the evaluation involves nondeterministic guessing.
	However, as the external atoms use only constant input, they are independent of the assignment are thus trivially \emph{monotonic} and \emph{antimonotonic}.
	In this case, the techniques from Section~\ref{sec:scalability}
	assign the correct truth value permanently after the first evaluation and thus this application
	is expected to benefit significantly from the improvements.
	
	\leanparagraph{AI in computer games}
	\emph{Angry-HEX} is an AI agent for the computer game \emph{Angry Birds}\footnote{\url{https://www.angrybirds.com}}~\cite{icghrstw2016-ieee}
	and was developed since 2012 for participation in the \emph{AIBirds competition}\footnote{\url{https://aibirds.org}};
	it was a finalist in 2015.
	The goal is to shoot birds with a slingshot at pigs located in buildings of wood, stone and ice blocks
	in order to destroy them. While the game is a strategy and skills game when playing manually, an AI agent can precisely
	compute the trajectory and the angle and speed in order to hit the desired target.
	Thus, the main issue is the selection of the best target.
	
	The strategy employed by Angry-HEX is to
	select the target which maximizes the estimated damage to pigs (primary goal) and to other objects (secondary goal).
	This is encoded as a \hex-program which
	guesses possible targets, estimates the damage for each, and uses weak constraints for optimization.
	However,
	the estimation of the damage requires physics simulation for deciding, for instance, which objects will fall if others
	are destroyed. As such a simulation cannot easily be done with rules alone, external atoms are used to interface
	with a physics simulator. Hence, the low-level simulation is done in external atoms while the high-level
	strategy is rule-based. The idea of this two-level approach is similar to the hybrid planning domain.

	\noindent \emph{Effects of improvements:}
	The application mainly benefits from the improvements in Section~\ref{sec:userconvenience}.
	It uses new language features from the ASP-Core-2 standard such as optimization statements.
	Moreover, until now a significant amount of development time was spent on low-level coding for interfacing physics libraries.
	The new Python interface is expected to speed up the development of the agent.
	Finally, the availability of binaries
	is more important than for other applications since the application needs to be run in an environment provided by the organizers of the competition.

\section{Conclusion}
\label{sec:discussion}

	The \dlvhex{} system implements \hex-programs and was first released ten years ago~\cite{eist2006b}.
	Over time, it was significantly extended with new algorithms, features, programming interfaces,
	and user's resources.
	While it served mainly as an experimental framework in the beginnings,
	its advancement towards practical applicability started only in the last three years.
	We now reached a stable state, where all extensions envisaged for this major release are implemented.

	In this paper, we gave a summary of version 2.5.0 and the most important recent enhancements.
	While literature on theoretical aspects and algorithms is preexisting, this paper focuses on the practical aspects
	which are relevant when realizing an application on top of \hex.
	After receiving positive feedback from individual users,
	we believe that informing the users succinctly about the enhancements will push the use not only of the new features
	but also of the system altogether.
	
	The improvements concern \emph{exploitation of known properties of external sources} for novel efficient evaluation algorithms
	and more flexibility of the language, and \emph{recent system extensions} for improved user's convenience;
	the latter include a Python programming interface, additional material and a new dissemination strategy.
	Real applications, which emerged independently of the research on \hex, but were realized on top of \dlvhex{},
	confirm the practicability of the approach.

\ifinlineref

\else
\bibliographystyle{acmtrans}
\bibliography{dlvhexprogressreport}
\fi

\ifextended

	\newpage
	\appendix
	\section{Selected Benchmark Results}
	\label{sec:benchmarks}

	In this appendix we discuss the performance effects of the presented system improvements on the applications from Section~\ref{sec:applications} in more detail.
	To this end, we include selected numerical results where experimental evaluation using benchmarks is appropriate.

	However, we recall that this paper does not only focus on efficiency improvements, but rather presents the current status of the system and application areas as a whole.
	Thus, while all presented applications benefit from some of the improvements (cf.~Section~\ref{sec:applications}), not all of them are suited to be used as performance benchmarks.
	Instead, the properties presented in Section~\ref{sec:externalsourceproperties} are also exploited for reducing syntactic limitations,
	and the improvements presented in Section~\ref{sec:userconvenience} concern usability of the system (i.e., ease its usage and the implementation of plugins).
	For an extensive evaluation solely focused on efficiency improvements we thus refer to~\citeN{efkrs2014-jair} and~\citeN{efkr2016-aij}.

	\leanparagraph{Hybrid planning}
	\citeN{Erdem2016a} considered two application domains from the robotics area. In the \emph{housekeeping domain}, multiple autonomous robots
	should tidy up a house within a given time limit by putting items to their places. Low-level feasibility checks concern the time limit and physical limitations of the robots.
	In the \emph{robotic manipulation domain}, two robots arrange elongated payloads, where payloads may only be moved by both robots together.
	Feasibility checks concern the avoiding of collisions of robots and payloads.
	
	Table~\ref{tab:hybrid} is an extraction from Table~5 by~\citeN{Erdem2016a}, containing the results which are in this context the most important ones.
	The authors used $20$ instances for each domain and present the
	average runtime needed to find the first feasible plan (\emph{first}) resp.~a maximum of $10.000$ feasible plans (\emph{max10k}).
	They compare multiple possibilities for implementing low-level feasibility checks, but in context of this paper only the two extreme cases are relevant, namely
	\emph{learning over complete interpretations} ({\sc Repl} by~\citeN{Erdem2016a})
	and
	\emph{learning over partial interpretations} ({\sc Int} by~\citeN{Erdem2016a}).
	It appears that the latter is significantly faster in both domains.
	
	\begin{table}[h]
		\centering
		\begin{tabular}[t]{l|rr|rr}
			\toprule
			& \multicolumn{2}{c|}{Housekeeping} & \multicolumn{2}{c}{Robotic Manipulation} \\
			learning method & first & max10k & first & max10k \\
			\midrule
			complete interpretations & $28865$ & $28893$ & $2000$ & $3748$ \\
			partial interpretations & $1019$ & $1112$ & $267$ & $1977$ \\
			\bottomrule
		\end{tabular}
		\caption{Hybrid planning, results in secs}
		\label{tab:hybrid}
	\end{table}

	Using the new version of \dlvhex{},
	the application would benefits from the possibility to provide three-valued oracle functions. While the developers had to hand-craft
	custom learning functions, which is a cumbersome task, the definition of a three-valued oracle function
	usually much more natural but allows for similar learning effects.


	\leanparagraph{Route planning}
	In this benchmark we consider route planning of a single person who	wants to visit multiple locations.
	If and only if the tour is longer than the given limit of $300$ cost units,
	the person wants to get refreshments.
	The external source allows only for computing shortest paths between two two locations.
	Thus, it cannot solve the task completely and there needs to be interaction with the \hex-program.
	
	The sequence in which the locations are visited is guessed non-deterministically in the logic program.
	While the direct connections between two locations are of minimum length by definition of the external atom,
	the length of the overall tour is only optimal wrt.~the chosen sequence of locations,
	but other sequences might lead to a shorter overall tour.
	However, we have the constraint that for visiting $n$ locations there should be at most $\lceil n \times 1.5 \rceil$ changes.
	Due to this constraint not all instances have a solution.
	The underlying map material is the public transport system of Vienna.
	The instance size is the number of locations $\#$.
	For each instance size $n$ we generated $50$ instances by randomly drawing $n$
	locations to visit.

	An implementation based on a full import of the map, i.e., \textbf{without exploiting the finiteness property}, yields \textbf{only timeouts} even for instances of the smallest size;
	we do not not explicitly shown in the table since there are only timeouts anyway.
	Table~\ref{tab:routeplanning-singlefull} shows the results if the \textbf{finiteness property and liberal safety} are exploited.
	It shows for each instance size the averages of the wall clock times, the grounding times, the solving times,
	the percentage of instances for which a \emph{solution} was found within the time limit,
	the average path \emph{length} (costs) of the instances with solutions,
	the average number of necessary \emph{changes},
	and the percentage of instances with solutions which include getting \emph{refreshments}.
	One can observe that for the considered instances with $1$ to $7$ locations (which is realistic wrt.~what is feasible within one day),
	the runtime is manageable.

	\begin{table}[t]
		\centering
		\setlength{\tabcolsep}{1.0mm}
		\begin{tabular}[t]{l|rrrrrrr}
			\toprule
			\# & \multicolumn{7}{c}{exploiting liberal safety based on finiteness property} \\
			& wall clock & ground & solve & solution (\%) & length & changes & refreshm. (\%) \\
			\midrule
			1 (50) & 2.40 ~~(0) & 1.71 ~~(0) & 0.54 ~~(0) & 100.00 & 0.00 & 0.00 & 0.00 \\
			2 (50) & 7.82 ~~(0) & 5.00 ~~(0) & 2.42 ~~(0) & 90.00 & 82.64 & 2.24 & 0.00 \\
			3 (50) & 16.44 ~~(0) & 9.46 ~~(0) & 5.81 ~~(0) & 76.00 & 152.21 & 3.92 & 0.00 \\
			4 (50) & 36.60 ~~(0) & 16.69 ~~(0) & 16.90 ~~(0) & 52.00 & 213.00 & 5.31 & 3.85 \\
			5 (50) & 102.71 ~~(0) & 26.63 ~~(0) & 69.26 ~~(0) & 52.00 & 281.27 & 7.58 & 11.54 \\
			6 (50) & 284.69 (38) & 236.43 (38) & 45.56 ~~(0) & 16.00 & 368.12 & 9.00 & 100.00 \\
			7 (50) & --- (50) & --- (50) & 0.00 ~~(0) & 0.00 & NaN & NaN & NaN \\
			\bottomrule
		\end{tabular}
		\caption{Route Planning benchmark, results in secs; timeout (``---'') is 300 secs}
		\label{tab:routeplanning-singlefull}
	\end{table}
\fi
	
\end{document}